\documentclass[10pt,twocolumn,letterpaper]{article}

\usepackage{cvpr}
\usepackage{times}
\usepackage{epsfig}
\usepackage{graphicx}
\usepackage{amsmath}
\usepackage{amssymb}
\usepackage{multirow}
\usepackage{graphicx}
\usepackage{booktabs} 
\usepackage{color}
\usepackage{bm}
\usepackage{bbm}
\usepackage{etoolbox}
\makeatletter
\patchcmd{\maketitle}
 {\def\@makefnmark}
 {\def\@makefnmark{}\def\useless@macro}
 {}{}
\makeatother

\usepackage[pagebackref,breaklinks,colorlinks]{hyperref}

\cvprfinalcopy % *** Uncomment this line for the final submission

\def\cvprPaperID{****} % *** Enter the CVPR Paper ID here

\begin{document}

 \author{Jiayi Shao, Xiaohan Wang, Ruijie Quan, Yi Yang \\
 \texttt{\small shaojiayi1@zju.edu.cn, wxh1996111@gmail.com, quanruij@hotmail.com, yangyics@zju.edu.cn} \\
ReLER Lab, CCAI, Zhejiang University
 }

\title{Action Sensitivity Learning for the Ego4D Episodic Memory Challenge 2023}

\maketitle
%\thispagestyle{empty}

%%%%%%%%% ABSTRACT
\begin{abstract}
In this report, we present ReLER's submission to two tracks in the Ego4D Episodic Memory Benchmark@CVPR 2023, including Natural Language Queries and Moment Queries. This solution inherits from our proposed Action Sensitivity Learning framework (ASL)~\cite{asl} to better capture discrepant information of frames. Further, we incorporate a series of stronger video features and fusion strategies. Our method achieves an average $m$AP of 29.34, ranking \textbf{1-st} in Moment Queries Challenge, and garners 19.79 mean R@1, ranking \textbf{2-nd} in Natural Language Queries Challenge. Our code will be released in \href{https://github.com/JonnyS1226/ego4d_asl}{https://github.com/JonnyS1226/ego4d\_asl}.
\end{abstract}

\section{Introduction}
Given an untrimmed egocentric video, the Ego4D~\cite{grauman2022ego4d} Moment Queries (MQ) Task aims to directly recognize actions from pre-defined categories and locate these action instances temporarily, while Ego4D Natural Language Queries (NLQ) task additionally incorporates a text query and seeks to locate the video segment revealing the answer to the query. These two tasks share the similarity that they both require locating the start frame and end frame, forming the time segment within the video. However, frames inside the segment are not equally valuable. We introduce action sensitivity to measure the importance of each frame. Some frames depicting the intrinsic content of actions are more sensitive to recognizing actions while others depicting the onset and offset are more sensitive to locating instances. Moreover, some transitional or blurred frames should be paid less attention to. In a word, precise MQ and NLQ solutions require capturing fine-grained information better. 

To this end, we leverge an Action sensitivity Learning framework~\cite{asl} based on a multi-scale transformer localization model to assess the action sensitivity of each frame. Then we utilize the generated action sensitivity to recalibrate the training process, guiding the model to focus more on sensitive frames. Combined with stronger pre-extracted video features and fusion strategies, our method outperforms all entries in Moment Queries Challenge and achieves the best performance on the test sets. Meanwhile, our method also ranked 2-nd on the public leaderboard of Natural Language Queries Challenge.

\begin{figure*}[t]
\vspace{-4mm}
\center
\includegraphics[width=1\linewidth]{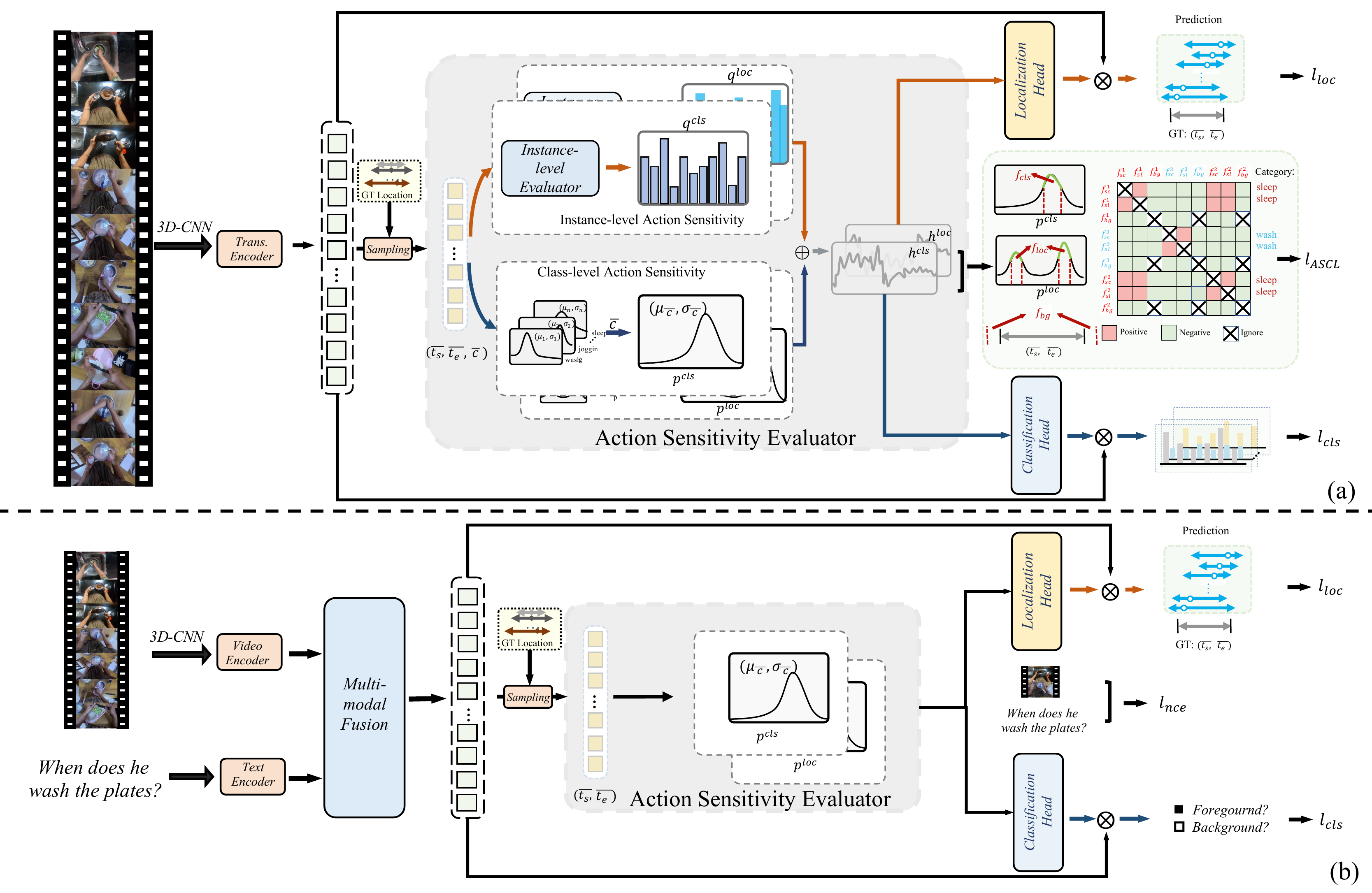}
\caption{
The overall framework of our approach. (a) depicts our approach to Moment Queries Challenge. (b) shows our approach to Natural Language Queries Challenge. Both are based on Action Sensitivity Learning framework~\cite{asl}.
}
\vspace{-2mm}
\label{fig:framework}  
\end{figure*}

%------------------------------------------------------------------------
\section{Related Work}
Ego4D Episodic Memory~\cite{grauman2022ego4d} is a large-scale egocentric video benchmark. The goal of Moment Queries Challenge is similar to Temporal Action Localization task. Unlike egocentric action recognition~\cite{Wang2021InteractivePL}, in this field, many previous works~\cite{lin2019bmn,gtad} follow the two-stage paradigm, generating action proposals first, then classifying and localizing these proposals, while others~\cite{zhang2022actionformer,afsd} are one-stage methods that detect actions directly. Our method falls into this part. Natural Language Queries Challenge can be regarded as Moment Retrieval or Video Grounding task, where many methods~\cite{vslnet,2d-tan} explore elaborated video-language interactions inspired by methods~\cite{wang2022alignandtell,zhao2022centerclip} to other multi-modal downstream tasks or are retrieve moments in a query-based manner~\cite{dformer_detr,momentdetr}. Besides, strong pre-trained video representations or features are of great benefits for action understanding, e.g., CLIP~\cite{clip}, VideoMAE~\cite{chen2022internvideo}, EgoVLP~\cite{egovlp}, where we explore some of these in the final submission.

\section{Methodology}
\subsection{Action Sensitivity Learning}
Moment Queries Challenge requires directly outputting a set of predicted action instances with categories and temporal boundaries, given a video clip. Natural Language Queries Challenge requires, given video clip and language query, outputting a temporal contiguous set of frames that answer this query. Our solutions to these two tracks all involve a novel action sensitivity learning framework~\cite{asl}.  The motivation behind it is that different frames have different importance to localization and classification. We introduce action sensitivity to measure the importance of each frame to sub-tasks (i.e. localization and classification). We utilize learnable class-aware Gaussian weights to model action sensitivity ($p^{cls}$ for classification and $p^{loc}$ for localization ). Then we apply it as loss weights of each frame to recalibrate training process. More details can be found in~\cite{asl}. For Natural Language Queries Challenge, we extend our proposed ASL framework to a multimodal version. The frameworks for two tracks are shown in Table~\ref{fig:framework}. We now introduce our specific pipelines for Moment Queries Challenge and Natural Language Queries Challenge.

\begin{table*}[]
    \small
    \centering
    \caption{\textbf{Results on the val set of Moment Queries Challenge:} SF and OV denote Slowfast~\cite{feichtenhofer2019slowfast} and Omnivore~\cite{girdhar2022omnivore} features. The best results are in \textbf{bold}. }
    \begin{tabular}{l l |c c }
        \toprule
        Method &  Feature &  Average mAP & Recall@1x, tIoU=0.5 \\
        \midrule
        Actionformer~\cite{zhang2022actionformer} & EgoVLP~\cite{egovlp} & 20.60 & 37.12 \\
        Actionformer~\cite{zhang2022actionformer} & SF + OV + EgoVLP~\cite{egovlp} & 21.40 & 38.73 \\
        InternVideo~\cite{chen2022internvideo} & InternVideo~\cite{chen2022internvideo} & 23.59 & 41.13 \\

        \midrule
        Ours base & EgoVLP~\cite{egovlp} & 20.79 & 39.26 \\
        Ours base & SF + OV + EgoVLP~\cite{egovlp} & 22.02 & 40.12 \\
        Ours base + ASL & EgoVLP~\cite{egovlp} & 22.83 & 40.67 \\
        Ours base + ASL & SF + OV + EgoVLP~\cite{egovlp} & 24.15 & 41.49 \\
        Ours base + ASL & InternVideo~\cite{chen2022internvideo} + EgoVLP~\cite{egovlp} & \textbf{27.85} & \textbf{46.98} \\
        \bottomrule
    \end{tabular}
    \vspace{0.2cm} 
    \label{apptab:mq_val}
    \vspace{-2em}
\end{table*}

\begin{table}[]
    \small
    \centering
    \caption{\textbf{Results on the test set of Moment Queries Challenge.}}
    \begin{tabular}{c |c c}
        \toprule
        Entry & Average mAP & Recall@1x, tIoU=0.5\\
        \midrule
        InternVideo~\cite{chen2022internvideo}  & 23.99& 44.24 \\
        mzs  & 26.62& 45.69 \\
        asl-ego4d(ours) & 29.34 & 48.50 \\
        \bottomrule
    \end{tabular}
    \label{apptab:mq_test}
    \vspace{-1.5em}
\end{table}

\begin{table*}[]
    \small
    \centering
    \caption{\textbf{Results on the val set of Natural Language Queries Challenge:} SF and OV denote Slowfast~\cite{feichtenhofer2019slowfast} and Omnivore~\cite{girdhar2022omnivore} features. The best results are in \textbf{bold}. The prime metric is mean R@1.}
    \begin{tabular}{l l |c c c |c c}
        \toprule
        \multirow{2}{*}{Method} &  \multirow{2}{*}{Feature} &  \multicolumn{3}{c}{ R@1(\%)} & \multicolumn{2}{c}{R@5 (\%)}\\
        & & IoU=$0.3$ & IoU=$0.5$ & mean & IoU=$0.3$ & IoU=$0.5$\\
        \midrule
        
        2D-TAN~\cite{2d-tan} & SF & 5.04 & 2.02 & 3.53  & 12.89 & 5.88\\
        VSLNet~\cite{vslnet} & SF & 5.45 & 3.12 & 4.28 & 10.74 & 6.63\\
        ReLER~\cite{liu2022reler} &  SF + OV + CLIP~\cite{clip} & 11.33 & 7.05 & 9.19 & 14.77 & 8.98 \\
        Actionformer~\cite{actionformer_nlq} & SF + OV + EgoVLP~\cite{egovlp}+CLIP~\cite{clip} & 15.72 & 10.12 & 12.92 & 34.64 & 23.64 \\
        InternVideo~\cite{chen2022internvideo} & InternVideo~\cite{tongvideomae} & 15.64 & 10.17 & 12.91 & 24.78 & 18.30 \\
        
        \midrule
        Ours base & SF + OV + CLIP~\cite{clip} & 13.80 & 9.51 & 11.66  & 35.28 & 23.13 \\
        Ours base + ASL & SF + OV + CLIP~\cite{clip} & 14.79 & 9.98 & 12.39 & 35.13 & 23.55 \\
        Ours base + ASL & SF + OV + EgoVLP~\cite{egovlp}+CLIP~\cite{clip} & 16.93 & 11.36 & 14.14 & 35.77 & 23.49 \\
        Ours base + ASL & InternVideo~\cite{tongvideomae} + EgoVLP~\cite{egovlp} & \textbf{22.62} & \textbf{15.64} & \textbf{19.13} & \textbf{46.86} &\textbf{32.16} \\
        \bottomrule
    \end{tabular}
    \label{apptab:nlq_val}
    \vspace{-1.5em}
\end{table*}

\begin{table}[]
    \footnotesize
    \centering
    \caption{\textbf{Results on the test set of Natural Language Queries Challenge.} The prime metric is mean R@1.}
    \begin{tabular}{c |c c c |c c}
        \toprule
        \multirow{2}{*}{Entry} &    \multicolumn{3}{c}{ R@1(\%)} & \multicolumn{2}{c}{R@5 (\%)}\\
         & IoU=$0.3$ & IoU=$0.5$ & mean & IoU=$0.3$ & IoU=$0.5$\\
        \midrule
        NaQ~\cite{ramakrishnan2023naq} & 21.70 & 13.64 & 17.67 & 25.12 & 16.33 \\
        ego-env & 23.28 & 14.36 & 18.82 & 27.25 & 17.58 \\
        asl-nlq(ours) & 24.13 & 15.46 & 19.79 & 34.37 & 23.18 \\
        GroundNLQ & 25.67 & 18.18 & 21.93 & 42.06 & 29.80 \\
        \bottomrule
    \end{tabular}
    \vspace{-2em} 
    \label{apptab:nlq_test}
\end{table}

\subsection{Track1: Moment Queries}
\textbf{Input Representations:} In this track, we use Slowfast~\cite{feichtenhofer2019slowfast} and Omnivore~\cite{girdhar2022omnivore} features provided by Ego4D. Inspired by the success of video masked autoencoders pretraining~\cite{tongvideomae,chen2022internvideo} and egocentric video-language pre-training~\cite{egovlp} methods, we additionally include EgoVLP~\cite{egovlp} and InternVideo~\cite{chen2022internvideo} features. Our method learns an MLP on each feature to project features and reduce dimension. Then these dimension-reduced features are concatenated and fed into the video encoder.
\label{moment queries}

\textbf{Feature Encoder and Sub-task Heads:} For fused features, inherited from ASL~\cite{asl}, we exert a Transformer encoder and pyramid network to encode feature sequences into a multiscale representation. To enhance representation, in Transformer encoder we operate temporal attention and channel attention parallelly and then fuse these two outputs. Then we model the action sensitivity for classification and localization of each frame in action instances. Our solution falls into a dense method so that the feature sequences are then processed by two sub-task heads (i.e. classification head and localization head, composed of 1D temporal convolutions), to generate dense final predictions. Meanwhile, via our proposed ASL~\cite{asl}, we obtain the action sensitivity $h(\bar c) \! \in \! \mathbb{R}^{N_f}$ (disentangled to classification and localization sub-task: $h(\bar c) \rightarrow \{h^{cls}(\bar c),h^{loc}(\bar c)\}$). $h$ is further used in training. 

 \textbf{Loss Function:} For classification sub-task,  we employ a focal loss~\cite{lin2017focal} to classify each frame, along with action sensitivity for classification   $h^{cls}$ :
\vspace{-0.3em}
\begin{equation} \small \label{equ:lcls}
\mathcal{L}_{cls}= \frac{1}{N_{pos}} \sum_i (\mathbbm{1}_{in_i} h^{cls}_i(\bar c_i) \mathcal{L}_{\text{focal}_i} 
    + \mathbbm{1}_{bg_i} \mathcal{L}_{\text{focal}_i})
\end{equation}
the indicators $\mathbbm{1}_{in_i}, \mathbbm{1}_{bg_i}$ denote if the $i$-th frame is within one ground-truth action or if belongs to the background, $N_{pos}$ represents the number of frames within action segments, while $\bar c_i$ indicates the action category of the $i$-th frame.

For localization sub-task, we adopt a DIoU loss~\cite{zheng2020distance} applied on frames within ground-truth action instance, to regress offsets from current frames to boundaries, combined with action sensitivity for localization $h^{loc}$:
\vspace{-0.5em}
\begin{equation} \small \label{equ:lloc}
\mathcal{L}_{loc}= \frac{1}{N_{pos}} \sum_i (\mathbbm{1}_{in_i} h^{loc}_i(\bar c_i) \mathcal{L}_{\text{DIoU}_i})
\end{equation}

 \textbf{Ensemble strategy:} We trained our model with different hyperparameters and scales. We ensemble these models for better performance. The ensemble strategy is: for $i$-th model, we get the output logits for classification $O^i_{cls} \in \mathbb{R}^{T\times C}$ and logits for localization $O^i_{loc} \in \mathbb{R}^{T\times 2}$. Then we do mean pooling on these logits to get the final ensembled outputs $O_{cls}=meanpooling(\{O^i_{cls}\}_{i=1}^E), O_{loc}=meanpooling(\{O^i_{loc}\}_{i=1}^E)$, $E$ is the number of models ensembled.

\subsection{Track2: Natural Language Queries:}
 \textbf{Input Representations:} In this track, we also use EgoVLP~\cite{egovlp} and InternVideo~\cite{chen2022internvideo} features for videos. Followed by~\cite{liu2022reler}, we use CLIP (ViT-B/16)~\cite{clip} to get the additional CLIP visual feature. All these features are projected by an MLP and concatenated in the channel dimension. For text queries, we use CLIP (ViT-L/14)~\cite{clip} to extract token-wise text embeddings. 
 
 \textbf{Text and Video Encoder:} We use two unimodal encoders for video features and text features respectively. For visual features, our encoder adopts a similar design as the encoder in MQ track in~\ref{moment queries}. To encode text features, we utilize multiple Transformer layers consisting of a linear projection layer and self-attention layers.
 
 \textbf{Multimodal Fusion and Sub-task Heads:} For multimodal feature fusion, our solution is built on a stack of cross-attention layers. For each cross-attention layer, query is obtained from video features while key and value are obtained from the same text features. Natural Language Queries Challenge can also be decoupled into two sub-tasks: the first is to predict if a frame is in the final answers, the second is to localize the answer's temporal boundary. Therefore we also adopt two sub-task heads: i) a classification head outputting a binary score for each frame. ii) a localization head outputting two distances from this frame to start time and end time. The designing of heads is the same as~\ref{moment queries}. Since ground truths in this track do not involve action categories, we only use one-class Gaussian to model the action sensitivity of frames (also decoupled to classification and localization sub-tasks) and apply this to classification and localization losses. 

 \textbf{Loss Function:} Our loss function is similar to~\ref{moment queries}, using Focal loss to classify and DIoU loss to localize combined with learned action sensitivity. Besides, we also utilize a NCE loss~\cite{liu2022reler,actionformer_nlq}, where for a text query, we consider a frame positive if it is inside a ground-truth answer and consider a frame negative if it is background.

 \textbf{Ensemble strategy:} To further enhance the performance, we ensemble our model with ~\cite{ramakrishnan2023naq}. We re-trained NaQ with EgoVLP~\cite{egovlp} and InternVideo~\cite{chen2022internvideo} features, which output topK results with respective confidence scores. Our ASL-based solution also outputs topK results.  We sort these predictions according to confidence scores and take the top-5 as the final results. The experimental result shows the efficacy of ensembling.

% % %-------------------------------------------------------------------------

% \begin{figure}[t]
% \center
% \includegraphics[width=\columnwidth]{./pic/data_augmentation.pdf}
% \caption{
% Illustration of data augmentation. (a) shows how the variable-length sliding window sampling strategy (SW) works. (b) shows how the video splicing strategy (VS) works. (c) is a combination of these two data augmentations which leads to better performance.
% }
% \vspace{-2mm}
% \label{fig:data_argu}  
% \end{figure}

\section{Experiments}

\label{sec:Experiments}

\subsection{Implementation Details}
For our solution to Moment Queries Challenge, we upsample the input video length to 1024. EgoVLP, InternVideo, Slowfast, and Omnivore features are respectively projected to 256, 128, 128, and 512 dimensions. The model embedding dimension is set to 1024. The number of attention heads is set to 16. The number of FPN layers is set to 8. We use a mini-batch size of 2, an epoch of 10 and a learning rate of $1e^{-4}$ with cosine weight decay and 5-epoch warm-up training strategies. During inference, the initial dense predictions are compressed with SoftNMS~\cite{softnms} and remain 2000 final predictions for submission. When ensembling, we use 3 models (i.e. $E=3$).

For our solution to Natural Language Queries Challenge,  EgoVLP, InternVideo, and CLIP are respectively projected to 256, 512, and 256 dimensions. In the backbone, we use 16 attention heads along with 512 dimensions. In training, we use a mini-batch of 16 and a learning rate of $1e^{-3}$ also with warm-up and cosine weight decay strategies. During inference, we sort the predictions according to confidence scores after processing with SoftNMS~\cite{softnms}, leaving 5 predictions for final outputs and for ensembling.

All experiments are conducted on NVIDIA Tesla V100 GPU. For Moment Queries Challenge, we report mean average precision (mAP) at tIoU thresholds [0.1:0.1:0.5] and report their average, i.e., the average mAP. Following official rules, we also report Recall@1x at tIoU=0.5, where x stands for the number of ground-truth instances of an action category. For Natural Language Challenge, we report Recall@1 (R@1) and Recall@5 (R@5) at tIoU thresholds of 0.3 and 0.5. For all these two tracks, we train our model on the training set when reporting results on the validation set (Table~\ref{apptab:mq_val} and~\ref{apptab:nlq_val}) and train our model on the combination of training and validation set for final submission to the test server (Table~\ref{apptab:mq_test} and~\ref{apptab:nlq_test}).

\subsection{Results: Moment Queries }
The comparison results on the validation set are shown in Table~\ref{apptab:mq_val}. Ours base means utilizing a Actionformer-like~\cite{zhang2022actionformer} model, which forward videos into multi-scale Transformers and directly predict categories and boundaries. Ours base + ASL means utilizing Action Sensitivity Learning to recalibrate training process. Using the same features of Slowfast~\cite{feichtenhofer2019slowfast}, Omnivore~\cite{girdhar2022omnivore} and EgoVLP~\cite{egovlp}, our methods outperforms Actionformer~\cite{zhang2022actionformer} by 2.23 of average mAP. Combining strong features of InternVideo~\cite{chen2022internvideo} and EgoVLP~\cite{egovlp}, our methods totally gains 4.26 of average mAP and 5.85 of Recall@1x(tIoU=0.5) compared to the champion~\cite{chen2022internvideo} in Moment Queries Challenge@ECCV 2022. Besides, our proposed Action Sensitivity Learning~\cite{asl} contributes to a 2.13 improvement in average mAP, indicating efficacy. As shown in Table~\ref{apptab:mq_test}, our ensembled methods finally obtain 27.85 average mAP, surpassing last year's top-ranked method by 4.26 on average mAP and taking 1-st place on the leaderboard of this challenge. However, the improvement of Recall@1x(tIoU=0.5) is not as significant as that of average mAP.

\subsection{Results: Natural Language Queries}
In the validation set of Natural Language Queries Challenge, the results are shown in Table~\ref{apptab:nlq_val}. Compared to Actionformer~\cite{zhang2022actionformer} (Runner-up in last year's challenge), our method improves mean R@1 by 1.22 under fair comparison. Among all these features (Slowfast~\cite{feichtenhofer2019slowfast}, Omnivore~\cite{girdhar2022omnivore}, EgoVLP~\cite{egovlp}, InternVideo~\cite{chen2022internvideo}), a combination of EgoVLP and InternVideo yields the best results as these two features are extracted using large-scale in-domain pretraining models. Besides, Action Sensitivity Learning though aims to tackle the task of Temporal Action Localization, also boosts a gain of 0.73 on mean R@1. On the leaderboard of this challenge (Table~\ref{apptab:nlq_test}), our ensembled methods finally arrive at 19.79 mean R@1, 2.12 higher than baseline (NaQ~\cite{ramakrishnan2023naq}) and ranked second.

\label{sec:Experiments}

\section{Limitation and Discussion}
Our key contribution in this solution is to capture the discrepant action sensitivity of different frames and apply the action sensitivity as weighting for losses to recalibrate training. Combined with stronger features and ensemble strategies, our method achieves good results on both Moment Queries and Natural Language Queries. As for limitations and future direction especially for NLQ challenge, our solution does not consider the context of egocentric. Meanwhile, more elaborated multimodal fusion designing may be explored while our method just use simple cross-attention layers.

{\small
\bibliographystyle{ieee_fullname}
\bibliography{egbib}
}

\end{document}